\documentclass[letterpaper]{article} 
\usepackage{aaai2026}  
\usepackage{times}  
\usepackage{helvet}  
\usepackage{courier}  
\usepackage[hyphens]{url}  
\usepackage{graphicx} 
\usepackage{natbib}  
\usepackage{caption} 
\usepackage{multirow}
\usepackage{booktabs}
\usepackage{threeparttable}
\usepackage{subfigure}
\usepackage{amsmath}
\usepackage{amsfonts}
\usepackage{algorithm}
\usepackage{algorithmic}

\usepackage{threeparttable}

\usepackage{newfloat}
\usepackage{listings}
\DeclareCaptionStyle{ruled}{labelfont=normalfont,labelsep=colon,strut=off} 
\floatstyle{ruled}
\newfloat{listing}{tb}{lst}{}
\floatname{listing}{Listing}
\title{FFGAF-SNN: The Forward-Forward Based Gradient Approximation Free Training Framework for Spiking Neural Networks}
\author{
Changqing Xu,
Ziqiang Yang,
Yi Liu,
Xinfang Liao,
Guiqi Mo,
Hao Zeng,
Yintang Yang\\
}

\affiliations{School of Microelectronics, Xidian University, Xi’an, China\\
\{cqxu, zqyang, yiliu, xfliao, gqmo, hzeng, ytyang\}@xidian.edu.cn}

\usepackage{indentfirst}
\begin{document}
\maketitle

\begin{abstract}
Spiking Neural Networks (SNNs) offer a biologically plausible framework for energy-efficient neuromorphic computing. However, it is a challenge to train SNNs due to their non-differentiability, efficiently. Existing gradient approximation approaches frequently sacrifice accuracy and face deployment limitations on edge devices due to the substantial computational requirements of backpropagation. To address these challenges, we propose a Forward-Forward (FF) based gradient approximation-free training framework for Spiking Neural Networks, which treats spiking activations as black-box modules, thereby eliminating the need for gradient approximation while significantly reducing computational complexity. Furthermore, we introduce a class-aware complexity adaptation mechanism that dynamically optimizes the loss function based on inter-class difficulty metrics, enabling efficient allocation of network resources across different categories. Experimental results demonstrate that our proposed training framework achieves test accuracies of 99.58\%, 92.13\%, and 75.64\% on the MNIST, Fashion-MNIST, and CIFAR-10 datasets, respectively, surpassing all existing FF-based SNN approaches. Additionally, our proposed method exhibits significant advantages in terms of memory access and computational power consumption.


\end{abstract}

%
\section{Introduction}
Spiking Neural Networks excel in processing spatiotemporal data and dynamic information, and they effectively mimic the functions of biological neural systems. Their strong biological explainability, temporal sensitivity, event-driven computation, and high energy efficiency make them highly promising for applications in neuromorphic computing, embedded systems, and edge computing\cite{xu2024stcsnn,xu2023ultra,pei2023albsnn}. With the continuous development of neuromorphic hardware and learning algorithms, SNNs are expected to play an important role in more fields.

Training Spiking Neural Networks presents significant challenges due to their complex dynamics and non-differentiable spiking activities. Existing methods can be categorized into three main types. The first type is inspired by biological synaptic weight adjustments, such as Spike-Timing-Dependent Plasticity (STDP)\cite{diehl2015unsupervised}, which relies solely on local neuronal activity without global supervision, thereby limiting performance. The second type involves indirect supervised training, where an Artificial Neural Network (ANN) is trained first and then converted into an SNNs\cite{diehl2015fast}; however, this approach fails to support real-time online learning. The third type comprises gradient-based methods with surrogate gradients\cite{bohte2000spikeprop}, which introduce errors due to gradient approximation and are prone to issues like vanishing gradients, exploding gradients, high computational complexity, difficulty in edge deployment, and a lack of biological plausibility. These limitations have driven researchers to explore innovative training approaches, such as biologically inspired algorithms.


The Forward-Forward (FF) algorithm, proposed by Geoffrey Hinton \cite{hinton2022forward}, offers a novel approach to training artificial neural networks. It replaces the traditional backpropagation method with two forward passes: one for positive data and another for negative data. This allows each network layer to have its own objective function, enhancing training efficiency. The FF algorithm is highly biologically plausible, aligning closely with neural activation mechanisms in the brain and avoiding the biologically implausible precise error signals needed in backpropagation. Additionally, it is hardware-friendly, especially for low-power devices, as it eliminates the computational demands of backpropagation. It also supports distributed learning, making it suitable for large-scale models. However, the FF algorithm has some limitations. In some experiments, its generalization performance may not match that of backpropagation algorithms. Training speed can also be slower compared to backpropagation. Moreover, effective contrastive learning with the FF algorithm requires high-quality negative data, which may be challenging to obtain in specific scenarios.

This study introduces a novel algorithm that combines the Forward-Forward algorithm with Spiking Neural Networks to address the training challenges of SNNs. The proposed method leverages the FF algorithm's ability to incorporate black-box layers, allowing the spiking neuron layer to be treated as a black box. This method avoids differentiating non-differentiable spiking activation functions and effectively solves the problem of SNNs training. Additionally, an improved ReLU activation function is employed to minimize the error introduced by the spiking neuron black-box layer.
The new algorithm enhances training accuracy while maintaining a low computational complexity. It preserves the high biological plausibility of the FF algorithm and achieves the low power consumption characteristic of SNNs. By combining these approaches, this method offers an efficient and biologically plausible solution for training SNNs.

This paper presents an innovative forward-forward based gradient approximation-free training framework for spiking neural networks, featuring three key technical breakthroughs.
First, we leverage the FF algorithm's unique capability of accommodating black-box layers by integrating SNN neuronal activation layers as non-trainable modules within the network architecture. During training, the SNN layers are frozen and do not participate in learning; they serve purely as an encoding mechanism during forward inference.
Second, we develop a dynamically weighted loss function that adapts to image complexity levels. 
Third, we propose a novel network structure that incorporates channel-wise weighting mechanisms. Our module ensures numerical stability through output regularization while employing an enhanced ReLU function as input to the SNN neuronal activation layer, thereby significantly reducing approximation errors.
The following summarizes our main contributions:
\begin{itemize}
\item We explored integrating spiking neurons into models based on the forward-forward algorithm. Using the FF algorithm's ability to incorporate black-box layers, we inserted spiking neurons as black-box layers into networks. This approach eliminates the need to compute the gradient of the activation function of the spiking neurons.
\item We propose a dynamic channel allocation method that enhances the loss function by adaptively distributing network channels across different categories based on their distinguishing complexity. This approach utilizes inter-channel similarity relationships to intelligently balance feature representation while maximizing network capacity utilization. The resulting framework automatically allocates more resources to challenging categories while maintaining efficient feature sharing, leading to improved performance on complex classification tasks without additional computational overhead.
\item To better adapt to spiking neurons, we refined the ReLU function to reduce information loss in the black-box layer. We also included a normalization layer to constrain the numerical range and stabilize training. Furthermore, we added weights across channels to enhance feature extraction.
\item 
For the first time, we have implemented learning in spiking convolutional neural networks using the Forward-Forward algorithm.
Experimental results show that our proposed FF-based SNN training method outperforms existing FF-based SNN methods on the MNIST, FashionMNIST and CIFAR-10 datasets. It also demonstrates strong competitiveness in FF-trained ANN networks.
\end{itemize}



\section{Related Work}
\subsection{SNN Training}
Training spiking neural networks is notoriously challenging, and current research is centered around three broad paradigms.  
First, non-backpropagation algorithms sidestep weight transport.DRTP\cite{frenkel2021learning} feeds a frozen random projection of one-hot labels forward layer-by-layer, yielding ultra-low memory training ideal for edge devices.F3 \cite{flugel2023feed} rescales gradients with a temporally delayed error, cutting the BP performance gap by 96\% while unlocking parallel weight updates.aDFA \cite{zhang2024training} further blends random feedback with local spike-timing cues to adapt the feedback alignment to spike regimes.  Complementing these are biologically grounded plasticity rules: canonical and its supervised variant sym-STDP \cite{hao2020biologically} fuse synaptic scaling with dynamic thresholds, whereas deep-STDP convolutional networks SDNN\cite{kheradpisheh2018stdp} learn hierarchies in an unsupervised latency-coding framework; all rely solely on local activity and remain modest in accuracy.Second, transfer learning via ANN-to-SNN conversion copies a pre-trained ANN into an SNN by matching spiking rates to the original analog activations—fast and accurate, yet biologically implausible.  Third, Directly supervised learning: using gradient substitution versus backpropagation  \cite{taylor2022robust,wang2020neuromorphic}. Although gradient substortion has the potential to achieve high performance, training still faces problems such as gradient substitution introduction errors, storage requirements of massive intermediate variables, low biological rationality, difficulty in hardware implementation, and gradient explosion.
\subsection{FF Algorithm}
The Forward-Forward algorithm, an emerging training framework without backpropagation, has gained attention recently.Hinton proposed the original FF algorithm.The Symmetric backpropagation-free contrastive learning(SymBa) algorithm\cite{lee2023symba}, which optimizes FF algorithm convergence via balancing positive/negative losses and using intrinsic class patterns (ICPs).  The trifecta\cite{dooms2023trifecta} presented the “Trio” technique, enhancing FF algorithm performance in deep networks through symmetric loss functions, batch normalization, and overlapping local updates. Convolutional channel-wise competitive learning algorithm\cite{cwcomp} improved the FF algorithm by introducing channel competitive learning in convolutional neural networks, proposing a new hierarchical loss function and Convolutional Feature Separation Extractor (CFSE) blocks.  Layer collaboration in the forward-forward algorithm\cite{lorberbom2024layer} highlighted that layer-independent training in FF algorithms restricts network optimization and reduces inter-layer collaboration, enhancing inter-layer information flow and cooperation by considering global “goodness” for each layer's optimization.
\subsection{Forward-Forward Algorithm in SNN}
Applying FF algorithms to SNN training offers an effective alternative to traditional backpropagation. Contrastive Signal-Dependent Plasticity (CSDP)\cite{ororbia2024contrastive} scheme based on FF learning rules, introducing activation trajectories for SNN self-supervised learning, boosting performance and enabling parallel learning/inference execution.  
Terres designed an SNN training framework based on the Forward-Forward algorithm for robust out-of-distribution detection, formulating SNN-compatible “goodness” functions that achieve energy-efficient OoD detection\cite{terres2024robustness}.
Backpropagation-free Spiking Neural Networks with the Forward-Forward Algorithm\cite{ghader2025backpropagation} developed a FF algorithm-based SNN training framework using two forward passes for inter-layer local learning, improving computational efficiency and hardware compatibility while delivering superior performance across multiple datasets. In summary, FF algorithms enhance SNN training efficiency, biological plausibility, and energy efficiency, showing great application potential. However, existing methods still fail to resolve the non-differentiability of neuronal activation functions and have relatively low implementation accuracy.

\section{Methodology}
This section introduces a forward-forward based gradient approximation-free training framework for spiking neural networks. We first analyze dataset class similarities to guide channel allocation, then develop a similarity-aware loss function for dynamic resource adjustment. This section also explains how an improved ReLU function mitigates information loss between SNN neuron layers during transmission, and it details the key architectural components required to train convolutional spiking neural networks with the Forward-Forward algorithm, presenting the complete training and inference pipeline.
\subsection{Inter-class Similarity Analysis}
\begin{figure}[ht]
\centering
\includegraphics[width=0.3\textwidth]{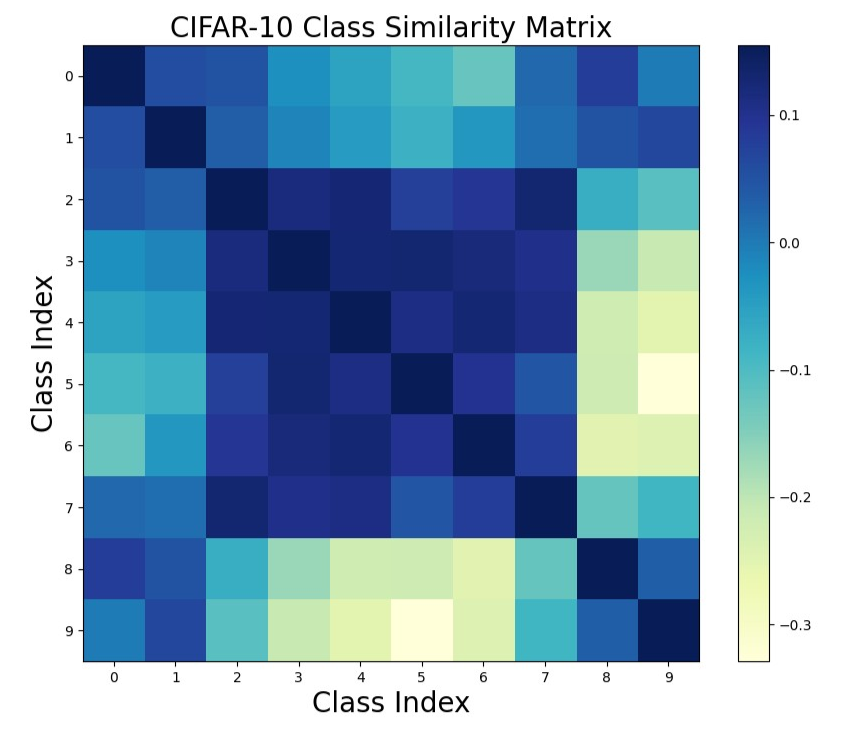}
\caption{Class similarity heatmap of the CIFAR-10 dataset, visualizing the mean pairwise feature similarities between all class combinations.}
\label{fig1a}
\end{figure}

\begin{figure}[t]
\centering
\includegraphics[width=0.4\textwidth]{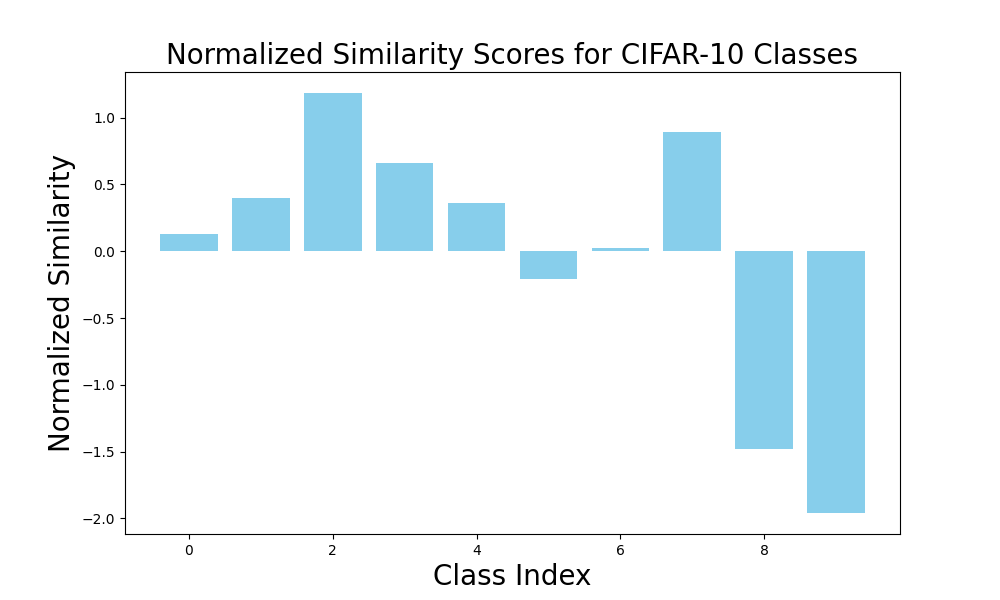}
\caption{The discriminative complexity of each class was quantified as the sum of its similarity measures to all other classes. The table presents normalized complexity scores for each category in the CIFAR-10 dataset.}
\label{fig1b}
\end{figure}

To calculate the inter-class similarity matrix, we start by extracting feature vectors from the dataset, where $\mathcal D$ denotes the number of datasets. For each sample $(x_i, y_i)\in\mathcal D$, \(\mathbf{f}_i\) is the feature vector of the \(i\)-th sample.  
The class-mean feature vector for class $c$ is computed as:
\begin{equation}
\mathbf{f}_c = \frac{1}{N_c} \sum_{i \in \{i \mid y_i = c\}} \mathbf{f}_i
\end{equation}
where \( N_c \) is the number of samples in class \( c \).

Next, we calculate the cosine similarity between each pair of classes to form the similarity matrix \( \mathbf{S} \):
\begin{equation}
\mathbf{S}_{c, c'} = \frac{\mathbf{f}_c \cdot \mathbf{f}_{c'}}{\|\mathbf{f}_c\| \|\mathbf{f}_{c'}\|}
\end{equation}
where \( \mathbf{f}_c \) and \( \mathbf{f}_{c'} \) are the average feature vectors of classes \( c \) and \( c' \), respectively. As depicted in Fig. 1, this matrix quantifies the similarity between classes based on their feature vectors. To analyze the overall similarity of each class, we compute the sum of similarities for each class \( c \) and normalize these sums to have zero mean and unit variance:
\begin{equation}
\tilde{s}_c = \frac{\sum_{c'=1}^K\mathbf{S}_{c, c'}-\mu_s}{\sigma_s}
\end{equation}
where \( K \) is the total number of classes, and \( \mu_s \) and \( \sigma_s \) are the mean and standard deviation of the similarity sums across all classes. As depicted in Fig. 2, this normalization step standardizes the similarity scores, allowing for a more interpretable comparison of class similarities.

To determine the channel allocation for each class, use the computed similarity \( \tilde{s}_c \) as the proportion for distributing channels reasonably.
 We incorporate the similarity metric $\tilde{s}_c$ as dynamic weighting coefficients to achieve an intelligent allocation of network channel resources. The innovation of this method is reflected in:
\begin{equation}
S_j = \left\lfloor \frac{\tilde{s}_j-\min{\tilde{s}_j}+\phi}{\sum_{c=1}^K(\tilde{s}_c-\min{\tilde{s}_c}+\phi)} \cdot C_{total} \right\rfloor
\end{equation}
where $S_j$ denotes the number of channels allocated to class $j$, and $C_{total}$ represents the total number of available channels. The hyperparameter \(\phi\), governing the uniformity of resource allocation, is fixed at 2 in this study. This proportion-based allocation mechanism ensures both the specificity of feature extraction and the maintenance of shared representation capabilities within the network.
\subsection{Loss Function}
To evaluate the performance of convolutional layers, we assume there are \( J \) data classes. The feature map activation outputs of each convolutional layer are denoted as \( \mathbf{Y} \in \mathbb{R}^{N \times T \times C \times H \times W} \), where \( N \) is the number of samples in the mini-batch, $T$ is the temporal dimension,\( C \) is the channel dimension, and \( H \), \( W \) are the spatial dimensions. 
We refer to \cite{cwcomp}, assuming \( J \) data classes, \( \mathbf{Y} \) is partitioned into \( J \) subsets, with the number of channels in each subset determined by the complexity of the corresponding class \( j \).
Thus, \( \hat{\mathbf{Y}}_j \in \mathbb{R}^{N \times T \times S_j \times H \times W} \) represents the activation output of  \( j^{th} \) class, where \( S_j \) is the number of channels allocated to class \( j \). The holistic goodness factor for each layer is a matrix \( \mathbf{G} \in \mathbb{R}^{N \times J} \). Each element \( G_{n,j} \) in this matrix quantifies the goodness of class \( j \) for example \( n \), computed via the spatiotemporal and channel-wise mean of the square of \( \hat{\mathbf{Y}}_j \), as shown below.

\begin{equation}
G_{n,j} = \frac{1}{S_j \times H \times W} \sum_{s=1}^{S_j} \sum_{t=1}^{T} \sum_{h=1}^{H} \sum_{w=1}^{W} \hat{Y}_{n,j,t,s,h,w}^2
\end{equation}
The cross-entropy loss function is defined as:

\begin{equation}
\mathcal{L} = - \sum_{n} \sum_{j} y_{n,j} \log(G_{n,j})
\end{equation}
Where \( G_{n,j} \) represents the predicted value and \( y_{n,j} \) represents the true value.

For the encoding layer training block, a represents the pre-activation values obtained after convolution followed by batch normalization, with the ReLU function serving as the activation function. The activation process is defined as follows:
\begin{equation}
Y_{n,c,h,w} = \text{ReLU}\left(\mathbf{a}_{n, c, h, w}\right)
\end{equation}
For the hidden layer training block, let a denote the pre-activation values obtained after convolution followed by temporal normalization\cite{zheng2021going}. The ReLU activation function is applied as follows:
\begin{equation}
Y_{n,c,h,w} = \text{ReLU}\left( \sum_{t=1}^{T} \mathbf{a}_{t,n, c, h, w} \right)
\end{equation}



\subsection{Network}
\begin{figure*}[t]
\centering
\includegraphics[width=1.2\columnwidth]{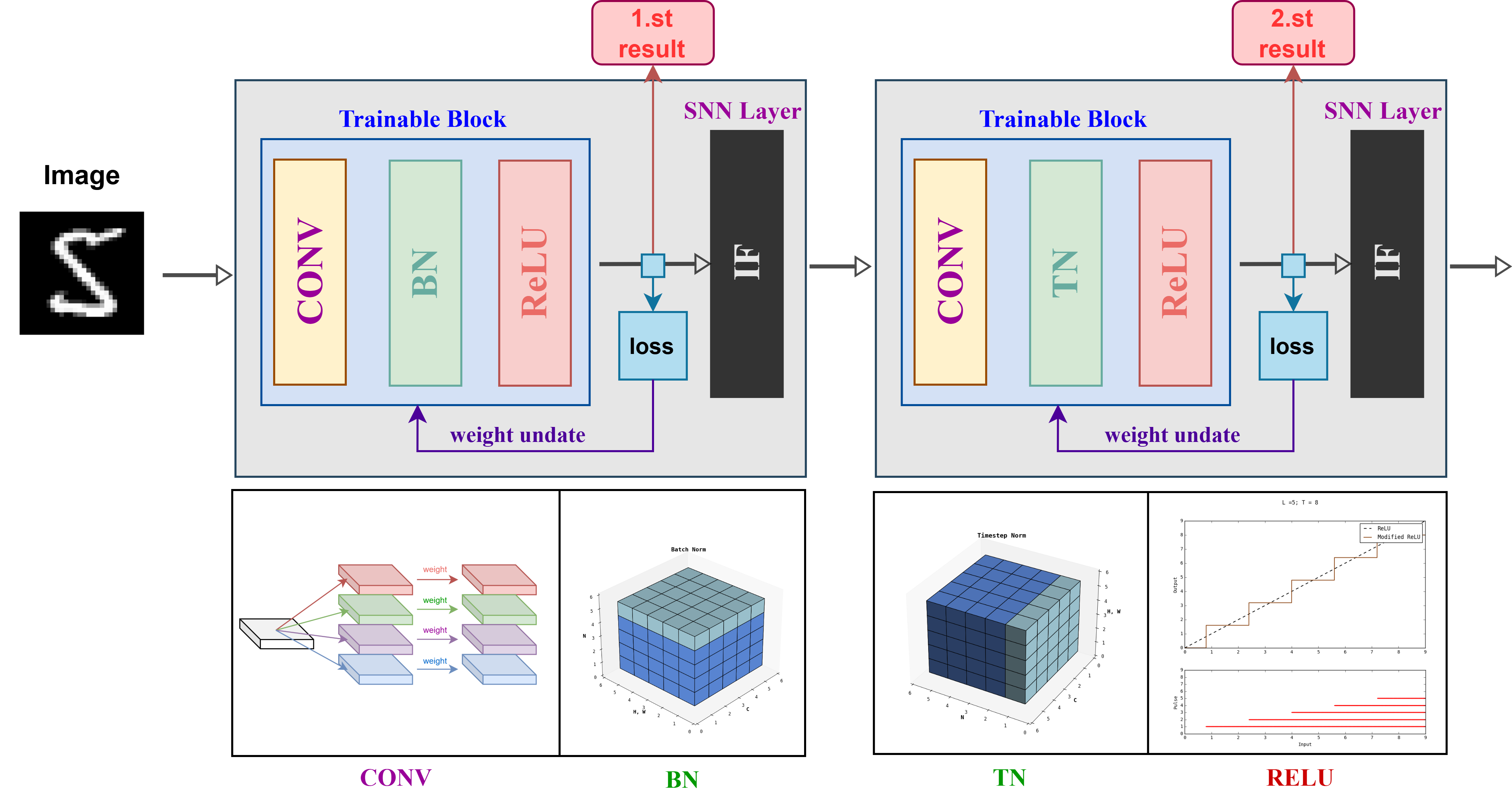}
\caption{SNN network architecture based on the Forward-Forward (FF) algorithm. The network comprises an input layer, convolutional layers, an SNN black-box layer that encodes inputs into spikes, and subsequent layers for feature extraction and classification. The SNN layer integrates with custom ReLU and regularization layers to facilitate training without directly deriving the SNN neuron activation functions.}
\label{fig1}
\end{figure*}

The network architecture consists of a trainable encoding block followed by three hidden blocks. The encoding block performs convolutional feature extraction on input images, applies batch normalization for stable activation ranges, and uses a custom ReLU activation function before feeding the outputs into the loss function. Each hidden block processes spiking inputs through temporal convolution and normalization, integrates features across time steps, and generates spike sequences via integrate-and-fire neurons receiving rectified inputs. This hybrid design combines the stability of normalized feature learning with biologically plausible spiking dynamics, while maintaining compatibility with gradient-based training through carefully designed activation pathways. The convolution process is illustrated as follows:
\begin{equation}
\hat{Y}_{t,n,j,h,w} = w_{j}(X \ast K_{j})_{t,n,h,w} + b_{j}
\end{equation}
Where \( X \) is {the input feature map},
\( w_{j}\) is the channel weight parameter,
\( K_{j} \) is the convolution kernel for channel \( j \),
\( \ast \) denotes the convolution operation,
\( b_{j} \) is the bias term for channel \( j \).
To constrain the output range, we regularize the output such that the resulting distribution has a mean of 0 and an adjustable standard deviation. This ensures the output lies within the expressible range of the SNN neuron layer and also stabilizes the training process. The $regularized output$ is scaled by a factor $thresh$, which controls the output range:
\begin{equation}
\text{$Regularized Output$} = \text{$thresh$} \times \frac{\text{$output$} - \mu}{\sigma}
\end{equation}
where \(\mu\) is the mean and $\sigma$ is the standard deviation of the output distribution. The parameter $thresh$ effectively adjusts the range of the regularized output to match the desired expressible range of the SNN neuron layer.

\textbf{Conversion error reduction:}
Given that our model incorporates an SNN neuron layer for information quantization and encoding, to ensure the completeness of information transmission within the network, we must analyze the transmission error.
We derive the potential update equation that describes the basic function of spiking neurons.

\begin{equation}
v^l (t) - v^l (t - 1) = W^l x^{l-1} (t) - s^l (t) \theta^l
\end{equation}

This equation represents the membrane potential update rule for the \( l \)-th layer neurons at each time step \( t \). The left-hand side \( v^l (t) - v^l (t - 1) \) indicates the change in membrane potential, the term \( W^l x^{l-1} (t) \) denotes the weighted input from the previous layer, and \( s^l (t) \theta^l \) represents the membrane potential reset of neuron models.

By summing this equation from time 1 to \( T \), we obtain:

\begin{equation}
v^l (T) - v^l (0) = W^l \sum_{i=1}^T x^{l-1} (i) - \sum_{i=1}^T s^l (i) \theta^l
\end{equation}

This equation shows the total potential change over \( T \) time steps for the \( l \)-th layer neurons. The left-hand side \( v^l (T) - v^l (0) \) reflects the overall potential change from the initial state to time \( T \). On the right-hand side, \( W^l \sum_{i=1}^T x^{l-1} (i) \) is the accumulated weighted input from the previous layer, and \( \sum_{i=1}^T s^l (i) \theta^l \) represents the accumulated impact of spiking on the potential reset.

To convert activation values into spikes using the formula, the conversion error \( v^l (T) - v^l (0) \) should ideally be zero. To reduce errors from the black-box layer encoding, we use a modified ReLU activation function,\cite{ann2snn}, enhancing training accuracy and efficiency.
The modified function helps in reducing the conversion error when transitioning between different quantization levels. The function is defined as follows:
\begin{equation}
\mathbf{a}^l = \widehat{h}(\mathbf{z}^l) = \lambda^l  \text{clip}\left( \frac{1}{L} \left\lfloor \frac{\mathbf{z}^l L}{\lambda^l} + \varphi \right\rfloor, 0, 1 \right).
\end{equation}
Here, \(\lambda^l\) is a scaling factor for layer \( l \), \( L \) is the number of quantization levels, and \(\varphi\) is a hyperparameter vector that controls the shift of the activation function. The improved ReLU function helps reduce the conversion error.

\begin{equation}
O_{t,n,c,h,w} = \text{SNN}\left( \mathbf{Y}_{n, c, h, w} \right)
\end{equation}
O represents the spiking output after passing through the SNN black-box layer.



\section{Experiments and results}

\begin{figure}[ht]
{
\centering
\subfigure[]{
\includegraphics[width=0.7\columnwidth]{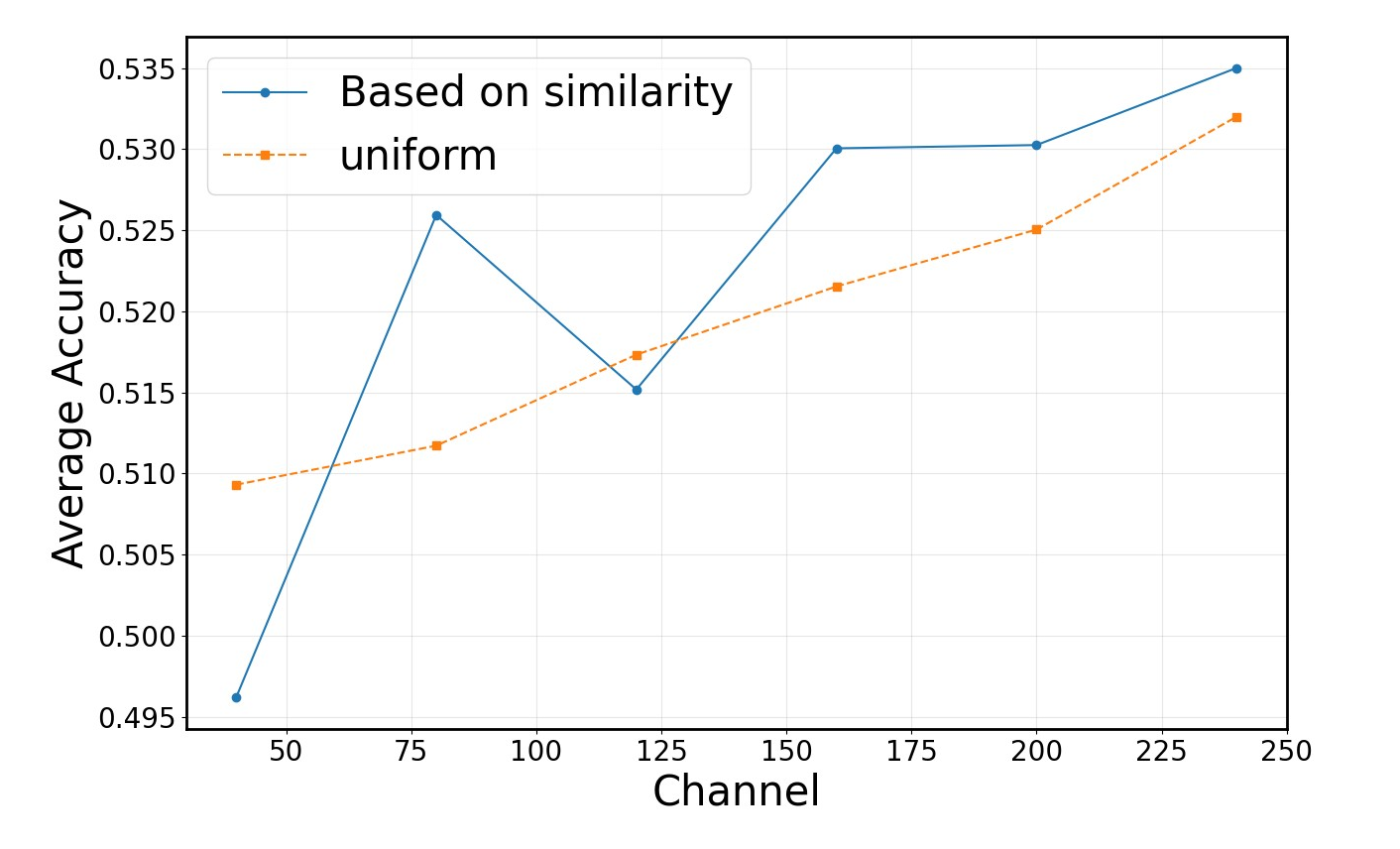}
\label{STEB}
}
\subfigure[]{
\includegraphics[width=0.7\columnwidth]{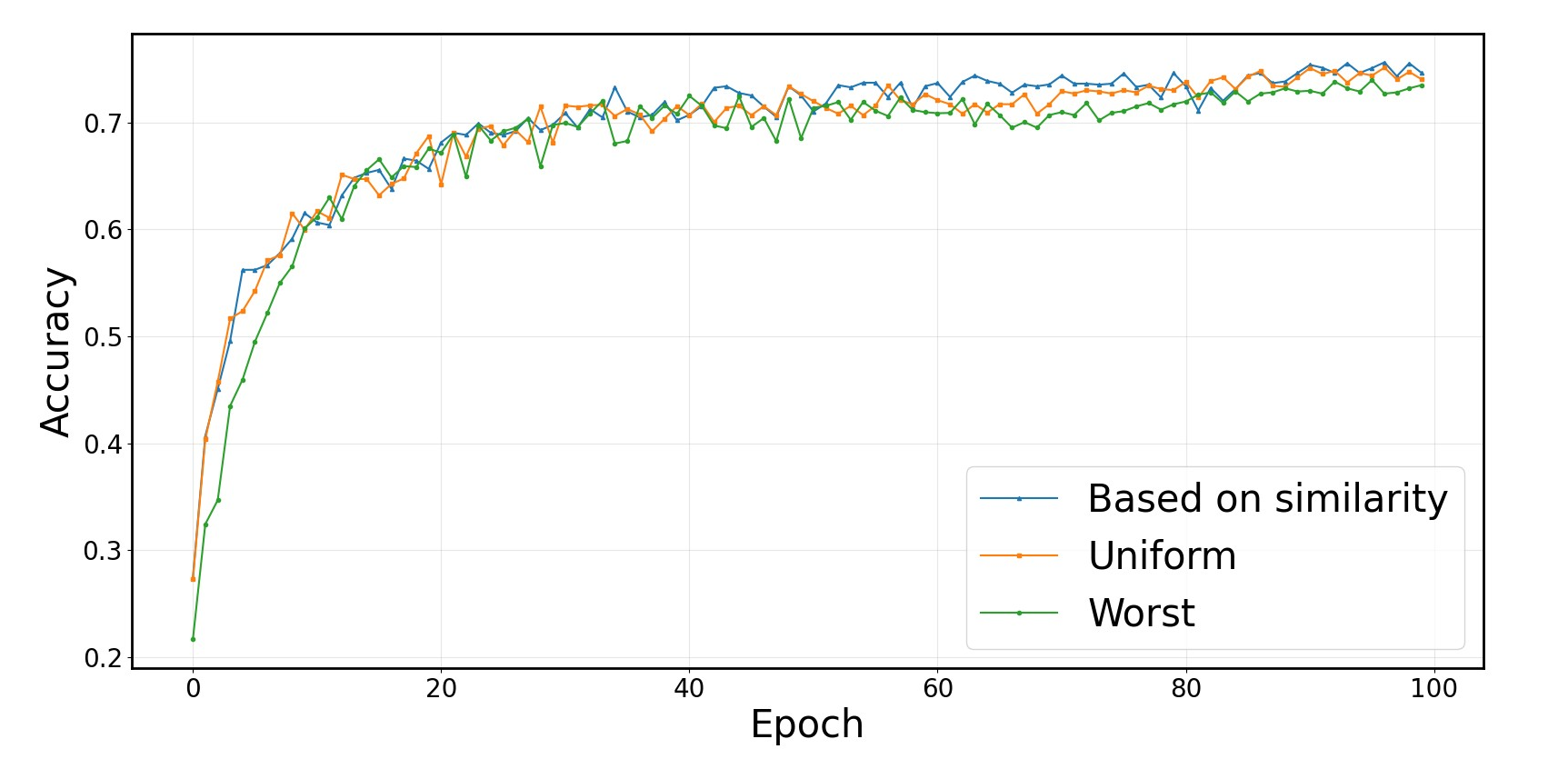}
\label{CAB}
}

\caption{(a)The figure demonstrates the accuracy variation of our network on the CIFAR-10 dataset under three different classification strategies: complexity-aware allocation, uniform allocation, and worst-case complexity-based allocation. (b)The figure presents the accuracy progression of a single-layer network on the Fashion-MNIST dataset as the number of channels increases.}
}
\end{figure}

This section details the experiments conducted to evaluate the proposed techniques and compare them with existing research. Model performance is assessed across three benchmark datasets commonly used in literature: MNIST \cite{mnist}, Fashion-MNIST \cite{fashion}, and CIFAR-10 \cite{cifar10}. The model was trained on an NVIDIA-TESLA V100 GPU for 30 epochs on MNIST,50 epochs on Fashion-MNIST, and for 100 epochs on CIFAR-10. The time step is configured to be 10.During the experiments, the learning rate for the convolutional layers was set to 0.01, with a batch size of 128. The convolutional stride was configured as [1, 2, 1, 2, 1], and the network architecture comprised layers with dimensions [40, 120, 120, 240].


\begin{figure*}[ht]
\centering
\includegraphics[width=0.8\linewidth]{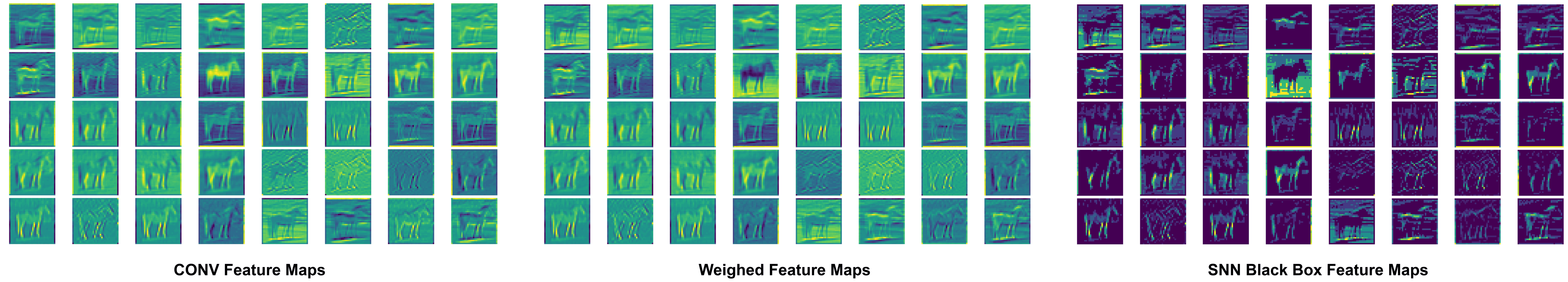}
\caption{CONV Feature Maps: Feature maps extracted by convolution, representing outputs across different channels.Weighted Feature Maps: Weighted output feature maps across different channels.SNN Black Box Feature Maps: Output from the SNN neuron layer, which serves as the input to the next layer.}
\label{fig5}
\end{figure*}

\subsection{Ablation Study}
Our ablation study demonstrates the impact of various model components, including our improved loss function, network architecture, weighted convolutions per channel, and activation functions. Experiments compare uniform channel classification (AV) and classification-aware-based (CAB) under the same network size, and analyze the effect of different layer architectures on performance.

The channel-competition-based loss function dynamically allocates resources according to discrimination complexity across different datasets, enabling efficient resource utilization. Specifically, more channels are allocated to hard-to-distinguish categories while fewer channels are assigned to easily separable ones. This strategy enhances model performance under constrained resources or reduces parameter requirements while maintaining comparable accuracy.
Experiments on the Fashion-MNIST dataset compare two allocation strategies in a single-layer network, analyzing accuracy trends with respect to parameter counts. Within the same network, we evaluate three allocation methods, complexity-aware allocation,  uniform allocation, and worst-case complexity-based allocation, demonstrating their impact on classification accuracy.

As illustrated in Fig. 4, our model achieves the highest accuracy when employing the complexity-aware classification strategy, followed by uniform classification, while the worst-case scenario (where allocation is inversely proportional to complexity) yields the lowest performance. It is noteworthy that since the CIFAR-10 dataset exhibits relatively slight inter-class discrimination difficulty variations, the advantage of our proposed method is not fully manifested - its effectiveness becomes more pronounced with datasets having greater inter-class disparity. In single-layer network comparisons between uniform and complexity-based classification, the proposed strategy demonstrates consistent superiority as channel numbers increase. However, when reaching the model's critical capacity point where resource constraints are alleviated, the accuracy gap between different strategies gradually diminishes and eventually converges. This phenomenon fundamentally validates our method's core advantage: intelligent complexity-aware resource allocation delivers superior performance, particularly under resource-constrained conditions.

Furthermore, we compare weighted and unweighted convolutions, where weighted convolutions incorporate trainable channel-wise importance weights to promote feature competition among channels, thereby improving inter-class discrimination. Additionally, we refine the ReLU activation function to mitigate errors introduced by the black-box layer.
Feature map visualizations across different network stages highlight the role of channel-wise weighting and the information propagation effects of the SNN black-box layer. As illustrated in Fig. 5, channel weighting selectively enhances discriminative features while suppressing less informative ones, whereas the SNN black-box layer effectively preserves key feature representations while filtering out noise.

\begin{table*}[ht]
\centering
\setlength{\tabcolsep}{10pt} 
\small 
\renewcommand{\arraystretch}{1} 
\begin{tabular}{@{}lcccccccc@{}}
\toprule
&&& \multicolumn{2}{c}{\textbf{CIFAR-10}} & \multicolumn{2}{c}{\textbf{MNIST}} & \multicolumn{2}{c}{\textbf{Fashion-MNIST}} \\
\textbf{Type}&\textbf{Method}&\textbf{Arch.}&\textbf{Test Er. (\%)} & \textbf{Epochs} & \textbf{Test Er. (\%)} & \textbf{Epochs} & \textbf{Test Er. (\%)} & \textbf{Epochs} \\
\midrule
\multirow{2}{*}{\textbf{BP}} 
 & Taylor et al. & SNN & NR & -& 97.91& 140 & 89.05& 140 \\
 & Wang Jing et al. & CSNN & NR& - & 99.0& - & NR & -\\
\midrule
\multirow{6}{*}{\textbf{Un BP}} 
 & FFF & CNN & 46.04& 200 & 97.16& 100 & NR& - \\
 & DRTP & MLP & 73.1 & 100& 98.98 & 100& NR& -\\
 & aDFA & SNN & NR& - & 98.01 & 20& 87.43& 20 \\
 &sym-STDP& SNN & NR & -& 96.73& 120 & 85.31& 120\\
 & SDNN & CSNN & NR& - & 98.4 & -& NR& - \\
\midrule
\multirow{7}{*}{\textbf{FF}} 
 & FF* & MLP & 54.89& 200 & 98.2 & 100& NR& - \\
 & PFF & RNN & NR& - & 98.66& - & 89.6& - \\
 & SymBa & MLP & 59.09 & 120& 98.58& 120 & NR& - \\
 & FFCL & MLP & 49.93& 50 & 97.23 & 50& 89.71& 50 \\
 & Layer Collaboration & MLP & 48.4 & 150& 98.8& 150 & 88.4& 150 \\
 & CaFo & CNN & 67.43& 5000 & 98.80& 5000 & NR& - \\
 & CwComp* & CNN & 75.28 & 100 & 99.27 & 30 & 91.79 & 50 \\
 & \textbf{FFGAF-SNN-ours} & \textbf{CSNN} & \textbf{75.64}& \textbf{100} & \textbf{99.58}& \textbf{30} & \textbf{92.13} & \textbf{50}\\
\midrule
\multirow{4}{*}{\textbf{FF-SNN}} 
 & Ororbia & SNN & NR & -& 97.58& 30 & NR& - \\
 & Ghader et al. & SNN & 54.03 & 300& 98.34& 300& 90.27& 300 \\
 & Terres et al. & SNN & 69.86& - & 97.77 & -& 85.75& - \\
 & \textbf{FFGAF-SNN-ours} & \textbf{CSNN} & \textbf{75.64}& \textbf{100} & \textbf{99.58}& \textbf{30} & \textbf{92.13}&\textbf{ 50} \\
\bottomrule
\end{tabular}
\captionof{table}{This table describes the test accuracy and training epochs of different training algorithms and network types across various datasets. FF* and CwComp* represent the accuracy achieved by our implemented models.}
\label{tab:table2}
\end{table*}

\begin{table*}[!ht]
\centering
\small 
\renewcommand{\arraystretch}{1} 
\setlength{\tabcolsep}{11pt}
\begin{tabular}{@{}lccccc@{}}
\toprule
\textbf{Model} & \textbf{Type} & \textbf{Parameters} & \textbf{Memory Access Energy(mJ)} & \textbf{Computational Energy(mJ)} & \textbf{Total Energy(mJ)} \\
\midrule
FF* & MLP & 18874368 & 0.157 & 0.059 & 0.216 \\
SymBa & MLP & 28311552 & 0.236 & 0.088 & 0.324 \\
Ghader et al. & SNN & 10144000 & 0.171 & 0.004 & 0.175 \\
Ororbia & SNN & 18874368 & 0.317 & 0.008 & 0.325 \\
CaFo & CNN & 627,768 & 0.012 & 1.993 & 2.005 \\
CwComp* & CNN & 588,133 & 0.065 & 0.390 & 0.455 \\
\textbf{CSNN-ours} & \textbf{CSNN} & \textbf{433,600} & \textbf{0.043} & \textbf{0.038} & \textbf{0.081} \\
\bottomrule
\end{tabular}
\captionof{table}{The following table compares the network parameters and theoretical computational energy consumption (including memory access and computation)\cite{lemaire2022analytical} of various models trained with the Forward-Forward (FF) algorithm for both Artificial Neural Networks (ANNs) and Spiking Neural Networks (SNNs).}
\label{tab:table1}
\end{table*}

\subsection{Comparison with Related Work}

We present comprehensive experimental results comparing our proposed Forward-Forward Gradient Approximation Free Training Framework (FFGAF-SNN)with state-of-the-art approaches. The FFGAF-SNN architecture employs an innovative design comprising one encoding layer training block and three hidden layer training blocks, each followed by an SNN IF neuron layer for output encoding and activation. To address feature information loss in the SNN neuron layer, we developed an enhanced ReLU function and incorporated channel-wise weight parameters for improved feature discrimination. The architecture utilizes batch dimension normalization in the first training block and temporal dimension regularization in hidden layers, demonstrating superior performance across MNIST, Fashion-MNIST, and CIFAR-10 datasets while significantly reducing model parameters. Notably, our framework achieves particularly remarkable results on CIFAR-10. It represents the first successful implementation of forward-forward training on CSNNs, offering substantial reductions in both parameter count and memory accesses compared to conventional FF-trained SNNs.
As shown in Table 1, the experimental results show that our method outperforms existing non-backpropagation learning algorithms, including those based on Direct Feedback Alignment (DFA-based) and Spike-Timing-Dependent Plasticity (STDP-based) approaches. Furthermore, our algorithmic model requires fewer iterations and converges faster than most models. Our model not only surpasses ANN models trained with DFA (FFF, DRTP), but also significantly exceeds the accuracy of Spiking Neural Network models trained with DFA (aDFA) by nearly 5\% on the Fashion-MNIST dataset. Compared with STDP-based SNN training methods (sym-STDP, SDNN), our approach also demonstrates higher accuracy. Testing on the Fashion-MIST dataset indicates that our model outperforms sym-STDP by nearly 7\%. Additionally, comparative studies with models trained using backpropagation show that our FFGAF-trained SNN can achieve comparable accuracy while eliminating the need for backpropagation (3\% higher on the Fashion-MINST dataset).
Additionally, our algorithm also outperformed FF-based ANN models such as PFF\cite{ororbia2023predictive}, FFCL\cite{karkehabadi2024ffcl}, CaFo\cite{zhang2024training}, and so on, demonstrating the superiority of our approach.

As shown in Table 2, a detailed energy efficiency analysis highlights the practical advantages of our FFGAF-SNN model. Compared to MLP networks trained with FF, this architecture maintains a significantly smaller parameter footprint—only 2.3\% of the parameter count of FF* models—while achieving approximately a 20\% improvement in accuracy on the CIFAR-10 dataset. Thanks to its fewer parameters, memory access energy consumption is substantially reduced relative to SNN models trained with FF. The memory access energy is only 14\% of the model proposed by Ororbia, and it remains favorable when compared to other CNN architectures. Although computational energy benefits from the inherent efficiency of SNN operations, the convolutional implementation results in slightly higher inference energy consumption than some SNN alternatives. However, compared to CNN models, our model's computational energy is only 9\% of CwComp* and our model's total energy consumption of 0.081mJ is significantly lower than other SNN models, which consume 0.175mJ and 0.325mJ. These combined advantages position our model as an exceptionally efficient solution for edge deployment scenarios.




\section{Discussions and Conclusion}

This paper presents the first successful implementation of the Forward-Forward (FF) algorithm for training convolutional spiking neural networks, introducing a novel FF Gradient Approximation Free (FFGAF-SNN) training methodology. Our innovative framework effectively combines the FF algorithm with SNNs by treating SNN neuron layers as black boxes, thereby elegantly addressing the non-differentiability challenge of spiking activities in conventional training while maintaining biological plausibility and reducing algorithmic complexity. We propose a category-complexity-based channel resource allocation strategy that optimizes the loss function for more efficient resource distribution compared to traditional approaches, particularly under parameter constraints. Drawing inspiration from ANN-to-SNN conversion, we develop an improved ReLU function to minimize information loss during feature map propagation through SNN layers, coupled with a channel-wise weighting mechanism that adaptively emphasizes essential features. Extensive experiments on MNIST, Fashion-MNIST, and CIFAR-10 datasets demonstrate that our approach achieves performance comparable to FF-trained ANNs while surpassing all existing FF-trained SNN models. The framework exhibits superior parameter efficiency and computational energy consumption compared to current FF-based methods. It shows competitive performance with traditional backpropagation algorithms across multiple datasets, highlighting its significant potential for practical applications in energy-efficient neural network implementations.



\bibliography{aaai2026.bib}
\end{document}